%% file: main.tex
\documentclass[article]{elsarticle}

\usepackage{color}
\usepackage{amsmath}
\usepackage{hyperref}
\usepackage{subfig}
\usepackage{adjustbox}
\usepackage[normalem]{ulem}

\definecolor{myblue}{RGB}{0, 128, 255}

\journal{arXiv}

\begin{document}

\begin{frontmatter}

\title{Digital Taxonomist: Identifying Plant Species in Community Scientists' Photographs}

\author[inst1]{Riccardo de Lutio}
\author[inst1]{Yihang She}
\author[inst1]{Stefano D'Aronco}
\author[inst1]{Stefania Russo}
\author[inst2]{Philipp Brun}
\author[inst1,inst3]{Jan D. Wegner}
\author[inst1]{Konrad Schindler}

\address[inst1]{EcoVision Lab, Photogrammetry and Remote Sensing, ETH Z\"urich}

\address[inst2]{Land Change Science, Dynamic Macroecology, WSL}

\address[inst3]{Institute for Computational Science, University of Zurich}


            

\begin{abstract}
Automatic identification of plant specimens from amateur photographs could improve species range maps, thus supporting ecosystems research as well as conservation efforts. However, classifying plant specimens based on image data alone is challenging: some species exhibit large variations in visual appearance, while at the same time different species are often visually similar; additionally, species observations follow a highly imbalanced, long-tailed distribution due to differences in abundance as well as observer biases. On the other hand, most species observations are accompanied by side information about the spatial, temporal and ecological context. Moreover, biological species are not an unordered list of classes but embedded in a hierarchical taxonomic structure. We propose a multimodal deep learning model that takes into account these additional cues in a unified framework. Our \emph{Digital Taxonomist} is able to identify plant species in photographs better than a classifier trained on the image content alone, the performance gained is over 6 percent points in terms of accuracy.

\end{abstract}

\begin{keyword}
Species Recognition \sep Community Science \sep Hierarchical Classification \sep Multimodal Learning
\end{keyword}

\end{frontmatter}

\input{intro}

\input{related}
 
\input{methodology}

\input{experiments}

\input{conclusion}
\newpage
\bibliography{main}

\input{appendix}

\end{document}

%% file: intro.tex
\section{Introduction}

Biodiversity describes the diversity of life in terms of species' numbers, similarity, abundance, and distribution across spatial scales \cite{barrotta2020,biodivbook}. Biodiversity is essential to human well-being but rapidly deteriorating worldwide in response to anthropogenic pressure~\cite{ipbes}. To effectively conserve biodiversity, its spatio-temporal distribution needs to be well understood, which requires efficient monitoring schemes.
Scientific surveys conducted at regional or country scales are, however, costly in terms of time and financial resources, as highly skilled professionals need to repeatedly examine extensive geographical areas and carefully document the encountered species. 

One viable way to complement professional biodiversity monitoring is the community science approach. The community science paradigm aims at involving the general public in scientific observations and investigations, and is particularly useful in cases where the \emph{experiment} is characterized by a large spatial and/or temporal scale~\cite{new_dawn_CS}. 
The community science approach has a long history in biodiversity monitoring \cite{eco_CS}. For example, volunteers have participated in the annual Christmas Bird Counts of the National Audubon Society in the USA since 1900 \cite{Butcher2007}.

With the rise of smartphones and other portable electronic devices, community science in biodiversity monitoring has  grown. Over the past decade, a multitude of smartphone apps have been released, allowing community scientists to conveniently report observations of plants and animals, and to upload images to online databases. Among the most popular of these apps is the 
iNaturalist~\cite{iNat_web} initiative, with over 3 million users and more than 36 million valid observations\footnote{A valid observation is an observation that has a date, a location, media evidence (image or sound), and has not been voted captive/cultivated.} distributed across the globe.

Although data gathered with community science is extremely valuable, it poses a number of challenges that need to be solved before it can be exploited effectively. One major issue is data quality, i.e., it is generally difficult to ensure that the collected data is correct and consistent. The main reasons are that community science data (either in the form of images or simple species presence observation) (\emph{i}) are collected by non-experts with varying training, expertise and skills, for instance, community scientists will on average not be able to name rare species as well as specialists; (\emph{ii}) often exhibit significant biases due to geographical variations in sampling effort, observation methods and traditions, as well as regional differences in infrastructure and accessibility.

In the context of biodiversity and species distribution mapping, Machine Learning (ML) can provide several tools for mitigating at least some of these limitations. For instance, the species recognition for the data collected on the field can be automatized to some extent to help the community scientist. This can either be done on-device to assist the user during data collection, as well as in a second step to assist the experts in verifying the user-supplied labels.
In recent years, computer vision has made great progress, mostly due to the rise of statistical ML. In fact, the application that spearheaded this development was the classification of image content into human-defined (semantic) categories~\cite{imagenet}. It is thus natural to ask whether ML can also assist community scientists to classify their photographs into taxonomic species, helping them to correctly identify what they have observed; thus paving the way towards more accurate and larger-scale species distribution maps.
Visual species recognition has been studied fairly extensively in recent years, with different image sources ranging from carefully collected zoological or botanical collections to uncontrolled outdoor and camera trap data~\cite{stanford_dogs,cub200,2020iwildcam}. In this paper, we specifically focus on the case of recognising plant species in data collected via community science applications such as iNaturalist~\cite{iNat_web} or Infoflora~\cite{infoflora_web}. Properties that distinguish this specific scenario from other image classification tasks include: (\emph{i}) Species observation numbers show an imbalanced distribution, as some species are naturally rare or harder to find and document than others (and perhaps also less attractive to photograph), such that they are rarely observed and only a few samples are available to train an ML model; (\emph{ii}) Side information is often readily available, e.g., the location and time when the image was taken are usually known, and in turn can be linked to further information like terrain maps, satellite images, etc.; (\emph{iii}) Biological species are related to each other in a hierarchical manner, i.e., through a taxonomic tree,%
\footnote{Namely, a sub-tree of the general hierarchy of (from top to bottom) kingdom, phylum, class, order, family, genus and species~\cite{taxoBook}.}
and one can leverage these relations during both training and inference. In particular, one may assume that, at any level of the hierarchy, species in the same group are, on average, more similar than species in distinct groups (see Fig.~\ref{fig:confusion}).

In this study, we develop an ML model for classifying community science photographs. Our focus is on how to best \emph{exploit side information} that comes with the actual photograph, to improve species recognition. 
By side information, we mean the locations and time points of the observations, as well as associated environmental variables and optical satellite imagery.
Location and time are usually uploaded together with the images.%
\footnote{These parameters constitute sensitive personal information, but community scientists are usually willing to disclose them to geo-locate their observations.}
Our model is inspired by other works such as~\cite{geo_aware,geo_prior}, however, there are a few key differences: (\emph{i}) we make use of additional metadata (altitude and Sentinel-2), (\emph{ii}) we train the model following a late fusion strategy and (\emph{iii}) we make use of the marginalisation loss  \cite{cloth_hier}.

Many environmental variables are publicly available, as are remote sensing images, e.g., the Sentinel-2 satellite data repositories~\cite{copernicus}.
Moreover, we include the taxonomic hierarchy to improve model performance at inference time.
Hierarchically structured class labels can be beneficial in two different ways: on the one hand, the hierarchy can be used as a regularisation of the model, which has been shown to improve the classification of rare classes~\cite{ozgur_crop}; on the other hand, the hierarchy can also be used at inference time to provide a prediction (at a coarser level) for species not present in the list of the output classes.
We investigate different strategies to exploit the side information and empirically compare them. We find that a model combining the community science images, spatio-temporal context, hierarchical labels and remote sensing images trained in a joint manner with a late fusion strategy performs the best. We validate the proposed method on a subset of the iNaturalist catalogue, with $56,608$ observations of $977$ distinct plant species, which includes observations of plant species across the territory of Switzerland.

%% file: related.tex
\section{Related Work}

\subsection{Context-based Modelling}
\label{sec:rel_geoprior}
Research has shown that the location context is important for modeling the distribution of species, and therefore can especially benefit fine-grained classification tasks.
In~\cite{rec_taxa} the authors adopt a nearest neighbour approach to predict the possible species that a person could encounter at certain locations given the previously recorded nearby observations. Although the paper acknowledges the fact that such information can be used to help and speed up species recognition, they do not combine their method with any image-based classification model.
In~\cite{birdsnap} the location and time where a photo was taken are used to define a prior distribution over bird species occurrences. An adaptive kernel density estimation is employed to construct that distribution, which is then combined with probabilistic output from a Support Vector Machine (SVM). Although the proposed method is effective when using spatial and temporal metadata to improve classification, the usage of SVM severely limited the overall performance. Novel, deep learning-based methods can achieve higher accuracies on the same dataset without spatio-temporal priors~\cite{birdsnap_best}.
With the fast advancement of deep learning, researchers have developed ways to utilise the location context with Convolutional Neural Networks (CNNs). In~\cite{loc_context} the authors investigate how to encode the image's GPS coordinate to increase prediction accuracy. The encoding is then concatenated with the image representation from the CNN before the final (linear) classifier. The paper also investigates the impact of further map features, e.g., precipitation maps, alongside simple GPS coordinates.
\cite{geo_aware} and~\cite{geo_prior} are two studies that combine deep learning and geographical information to improve species recognition accuracy.
In~\cite{geo_aware} the authors propose a refinement network that merges the prediction from a CNN with a secondary network that receives as input the location where the image was taken. The weights of the CNN network are kept frozen while training the refinement module. As a second option, the paper proposes a method where the location-aware network can alter the feature extraction inside the CNN, based on the picture's location. This second technique, however, did not lead to a substantial improvement.
\cite{geo_prior} propose a slightly different solution for the same problem, in this case the network responsible for extracting the geographical prior is in fact trained separately. The problem in this case is that the dataset consists exclusively of positive labels, i.e., it contains no information where the context speaks \emph{against} a certain species label. To overcome this, the authors propose a joint embedding loss able to deal with presence-only datasets.
The difference between the two approaches is that in the former~\cite{geo_aware} the geographical network is trained to improve the image-based prediction coming from the CNN, but cannot make a meaningful prediction on its own, i.e., without the CNN; whereas in the latter work~\cite{geo_prior} the geographical network is trained separately and can also be evaluated without an image, effectively producing a species distribution map.

\subsection{Hierarchical Labels}
Complementary to location context, structure among the species labels helps the classification task by sharing features among related (i.e., nearby) classes.
In~\cite{tree_prior}, the output classes are organised in a hierarchical structure, and features are transferred between related classes to inject the a-priori hierarchy into the deep neural network classifier. 
\cite{yan2015} was another early work that tackled hierarchical classification in the context of visual recognition. The proposed method is limited to a 2-level hierarchy, and it is composed of two classifiers: a coarser one, which separates more easily distinguishable classes, and a finer one the resolves the more difficult cases.
\cite{incremental1}, and more recently~\cite{incremental2} analysed the use of hierarchical labels for visual recognition for the specific case of incremental learning.
A hierarchical classifier for clothing recognition was proposed in~\cite{cloth_hier}. The model predicts a label hierarchy instead of a single label for the input object, by analyzing detection errors. The method exhibits good generalization capabilities also for novel clothing products that were not seen during training. 
In the past years, researchers have explored different ways to inject knowledge about hierarchical labels into neural networks. The authors of~\cite{fine_hier} propose a framework to predict the category scores at each hierarchy (tree) level in a top-down manner, with a multi-head network where each branch is responsible for a different level.
Recently,~\cite{dhall2020} have investigated and compared a number of strategies and loss functions to integrate hierarchical semantic structure into a CNN, including per-level classifiers, hierarchical $softmax$, and a marginalisation loss.
The marginalisation loss summarizes the hierarchical information in a bottom-up manner and, although being one of the simplest approaches, emerged as one of the most effective.
In~\cite{ozgur_crop} the authors investigate the task of classifying agricultural crops from a sequence of satellite images, where the crop labels also exhibit a hierarchical structure (e.g., wheat is more similar to other cereals than to, say, orchards). They propose a convolutional recurrent architecture, where increasing depth in the spatial/convolutional dimension corresponds to a finer hierarchy level, thus deriving higher-level features for finer classification from coarser lower-level features. The layout is specific to the recurrent setup and it is unclear how to adapt it to conventional CNNs without disrupting the feature extraction backbone.

As a general comment, we note that methods designed for hierarchical labels tend to use custom architectures and cannot easily be combined with well-known, pre-trained high-performance backbones.

%% file: methodology.tex
\section{Methodology}
\label{sec:methodology}

We now outline our proposed model for plant species classification. The model can be understood as composed of two branches: the first branch infers a probability distribution over plant species, by looking exclusively at the input image; the second branch infers another species distribution only from the auxiliary information, which is then combined with the image-based prediction to obtain a refined posterior distribution. The entire two-branch network is supervised jointly with a hierarchical loss that leverages the structure of the taxonomy. 

\begin{figure}[ht]
\begin{center}
\includegraphics[width=0.95\linewidth]{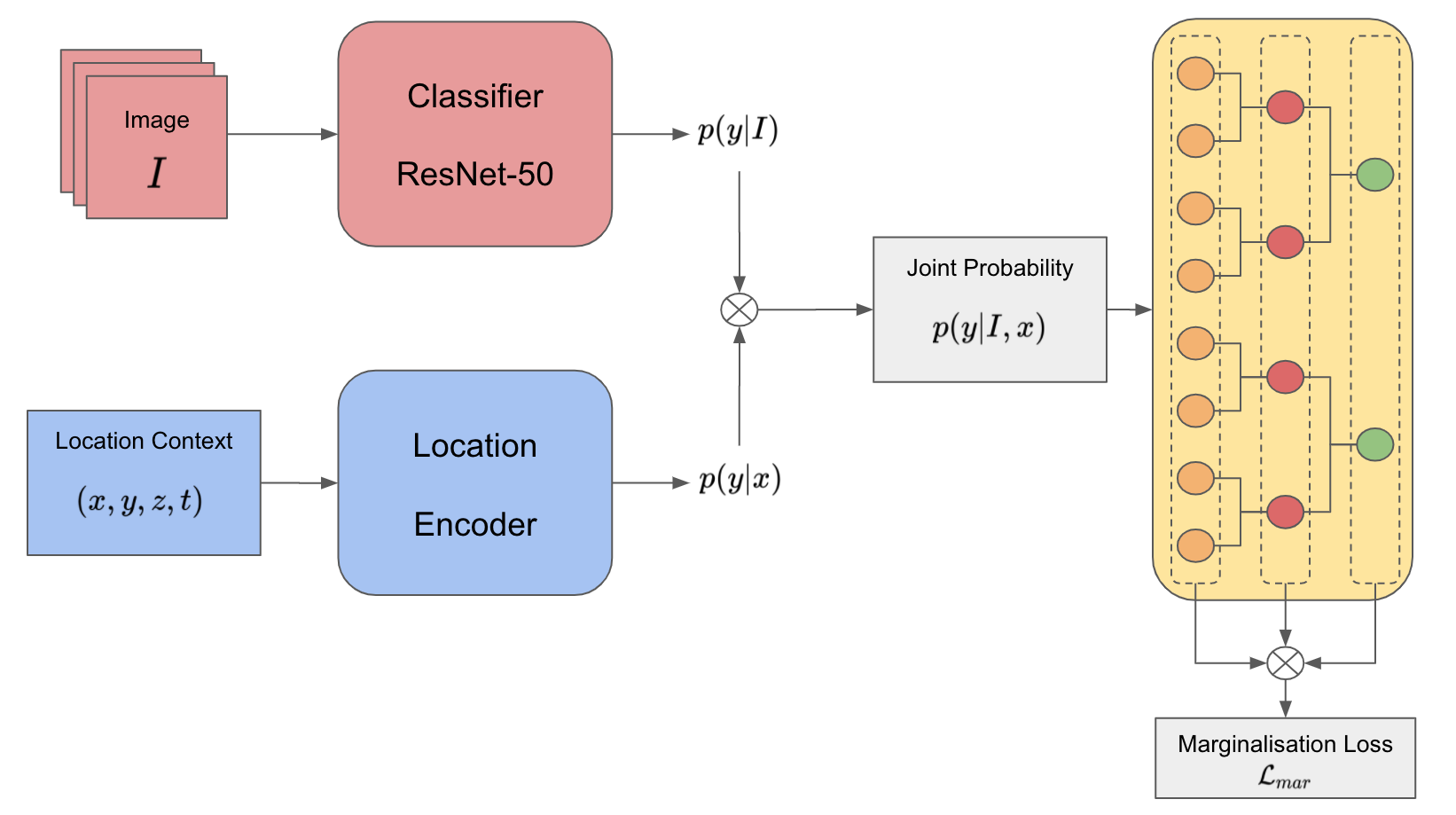}
\end{center}
   \caption{Overview of our model.}
\label{fig:model_overview}
\end{figure}

\subsection{Inference from Image}

Given an image $I$ that depicts a certain plant specimen, we can use a CNN to infer its species $y$. The network outputs a probability distribution $p(y|I;\theta)$ over all $C$ possible species, where $\theta$ are the learnable parameters (convolution weights). To lighten the notation we drop $\theta$ when it is clear from the context, and simply write $p(y|I)$.
In our implementation we use the popular ResNet architecture~\cite{resnet_paper}, although other networks could also be employed. Our ResNet is pre-trained on ImageNet~\cite{imagenet}, a setting that has become common practice to speed up training and boost performance with limited data.

\subsection{Inference from Spatio-temporal Context}

\begin{figure}[ht]
\begin{center}
\includegraphics[trim=3.5cm 0.cm 4cm 0.cm, clip, width=\linewidth]{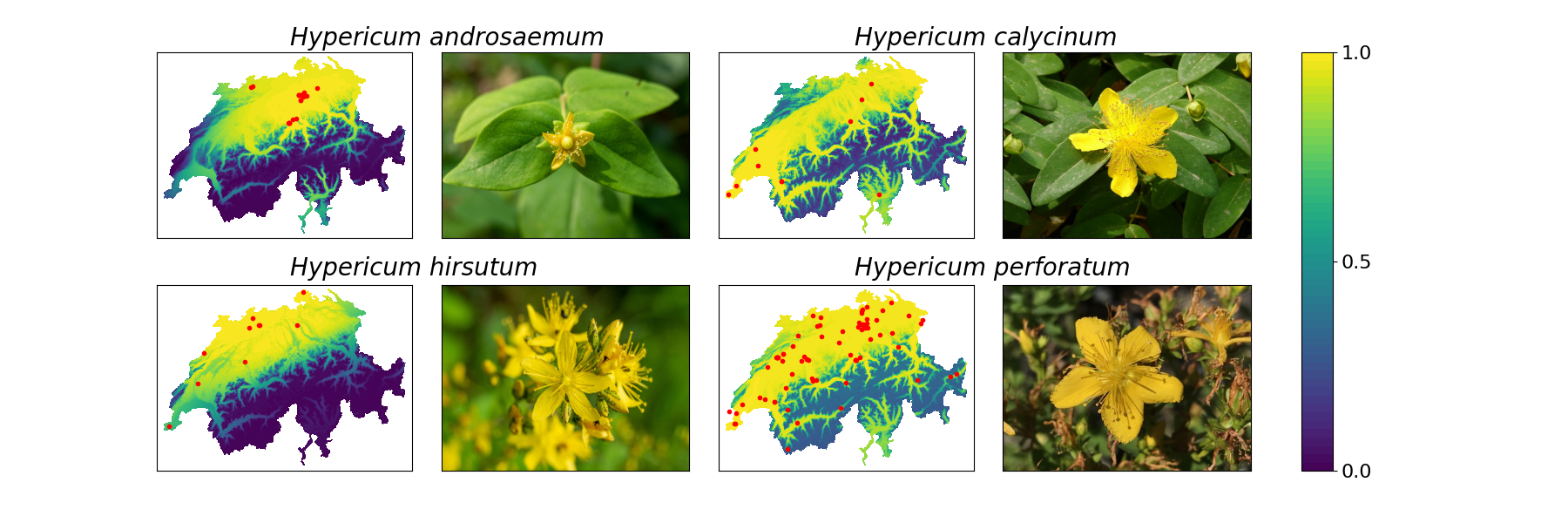}
\end{center}
\caption{Different \textit{Hypericum} species, in order \textit{H. androsaemum}, \textit{H. calycinum}, \textit{H. hirsutum} and \textit{H. perforatum}. The present species are visually similar but have different geographical distribution ranges. For such groups of species additional spatio-temporal information can help to improve classification accuracy. For each species we visualise the probability score learned by our Location Encoder (left), the location of the training samples (red dots) and a sample image from our training set (right).}
\label{fig:class-examples}
\end{figure}

As explained above, community science observations are often accompanied by auxiliary information, in particular spatio-temporal context, i.e, where and when the photo was taken.
We denote that spatio-temporal context by the vector $\mathbf{x}$. The spatial information includes longitude ($x$), latitude ($y$), and altitude ($z$), while the day of the year $t$ represents the temporal information.%
\footnote{Thus assuming the distribution is seasonally varying but stationary over a few years.} %
This information is typically included in the images' metadata, except for the altitude, which can be easily derived from the location given a Digital Elevation Model (DEM).
The spatio-temporal context of an observation has been shown to be a useful cue for classifying species observations (see Sec.~\ref{sec:rel_geoprior} and Fig.~\ref{fig:class-examples}) -- which is not surprising, as the probability of observing a certain species varies greatly across space and time.

Several methods have been proposed to merge such auxiliary information into the classification, for instance see~\cite{geo_aware,geo_prior}. We will briefly describe the different strategies and highlight their pros and cons:

\label{sec:train_strat}
\begin{description}
    \item[\textbf{Early Fusion}] In this case the image $I$ and auxiliary information $\mathbf{x}$ are together fed into a model which shall directly predict $p(y|I,\mathbf{x};\theta,\phi)$. That model is trained by minimizing a suitable loss function such as the cross-entropy between the predicted and true labels.
    The advantage of such an approach is that it does not impose any independence assumptions and the model can, in principle, leverage any statistical relation between $y$ and the inputs, including correlations between $I$ and $\mathbf{x})$. However, this generality comes at a price: (\emph{i}) at inference time the complete auxiliary information $\mathbf{x}$ must be fed to the model to obtain a reliable prediction, and (\emph{ii}) if the training data is scarce, processing the two sources $I$ and $\mathbf{x}$ together increases the risk of over-fitting to spurious correlations. 
    
    \item[\textbf{Separate Training}] This approach, exemplified by~\cite{geo_prior}, takes the opposite route and employs two completely separate networks: one ``main" network processes only the image to obtain $p(y|I;\theta)$, the second ``auxiliary" one processes only the side information to obtain $p(y|\mathbf{x};\phi)$. The two networks are trained separately and produce separate scores that are only merged at inference time. This corresponds to  the assumption that $I$ and $\mathbf{x}$ are independent, such that $p(y|I,\mathbf{x}) \propto p(y|I)\cdot p(y|\mathbf{x})$. The main advantage of this approach is a much reduced danger of over-fitting, as visual information and context are decorrelated. A further advantage is that one can use additional datasets without images to train the spatio-temporal prior. On the other hand, training that prior without supporting image information can also be difficult, particularly in the common situation with presence-only annotations~\cite{geo_prior}. Finally, any real correlations between $\mathbf{x}$ and $I$ will be lost, by construction.%
    \footnote{Such patterns are likely to exist. Examples include location-specific shadows or time-dependent snow cover.}
    
    \item[\textbf{Late Fusion}] This approach, employed for instance as one of the methods in~\cite{geo_aware}, constitutes a compromise between early fusion and the separate training.
    Separate branches are maintained for $I$ and $\mathbf{x}$. But their scores are not only combined during inference but also during training, with a joint loss function on the combined prediction $p(y|I,\mathbf{x})$. The risk of over-fitting remains low compared to early fusion, as the model admits correlations between visual and auxiliary cues only ``globally", but not between individual variables: $p(y|\mathbf{x})$ acts as a spatio-temporally varying rescaling of the image-based class scores $p(y|I)$, and vice versa. At the same time, presence-only observations do not challenge the training of the spatio-temporal prior, as the loss is computed only after including the visual information.
\end{description}

All the aforementioned methods are legitimate design choices, whether to prefer one or the other depends on the particular problem as well as the available data. In the experiment section, we empirically compare their performance for plant species classification.
In terms of network architecture, for separate training and late fusion, the auxiliary information is first embedded into a $C$-dimensional vector with a fully-connected network ($FCN_{\mathrm{context}}$), with $C$ the number of classes. The $FCN_{\mathrm{context}}$, with parameters $\phi$, has as last layer a sigmoid, such that its output represents a presence/absence probability \emph{per class}. Note that the sigmoid (rather than a softmax over $C$ classes) is chosen to reflect that, at a given place and time, multiple species can be present with high probability.

\subsection{Inference using Auxiliary Sentinel-2 Images}

Finally, given that we know the location where a specific species observation was made, we can extract additional context information from remotely-sensed sources, to potentially improve species identification performance. To illustrate this, we add a Sentinel-2 image of the region around $\mathbf{x}$ as further auxiliary data. Sentinel-2 was chosen for its potential to supplement meaningful information about the local ecosystem: it provides complete coverage of the region of interest (Switzerland). We choose to only use the 4 bands with the highest spatial resolution (10 m GSD) across the visible and infrared spectrum (ranging from 0.5 to 1.0 $\mu$m). These are commonly used to derive vegetation information and have been shown to be sufficient to derive further vegetation parameters \cite{nico_tree}.

The satellite data $S$ is fed into the model in a similar fashion as the location context. The only difference is that the embedding of the raw data into the $C$-dimensional vector $p(y|S;\psi)$ is a convolutional encoder with parameters $\psi$ (rather than a fully-connected network), to account for the nature of image data. In our implementation we use a ResNet-50.

As before, the embedded satellite imagery is combined with the other inputs according to the late fusion strategy and all three branches are trained jointly, via the merged score $p(y|I,\mathbf{x},S)$.

\subsection{Integration of Taxonomic Hierarchy}

\begin{figure}[ht]
\begin{center}
\includegraphics[width=0.5\linewidth]{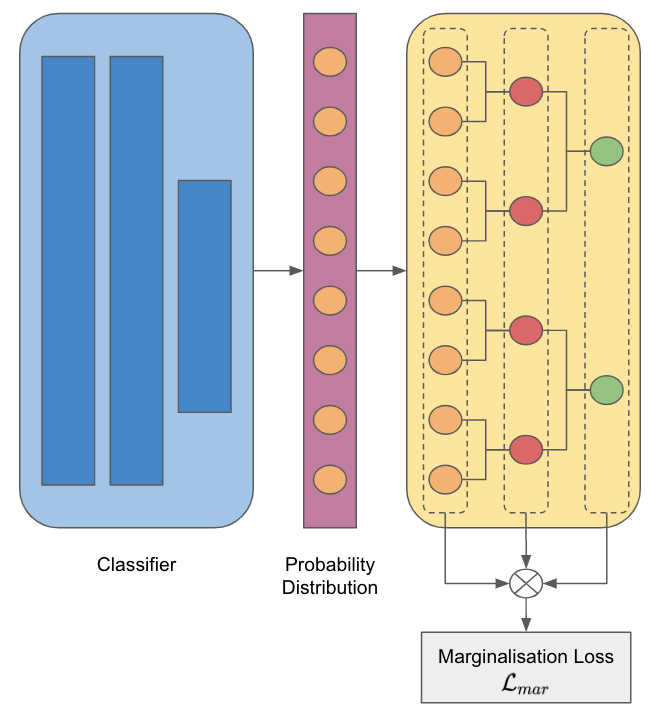}
\end{center}
   \caption{The idea of the marginalisation loss is to simultaneously apply a cross-entropy loss at all levels of the taxonomic hierarchy. As the output of the classifier is the probability distribution over all species, marginalising over all species within each genus yields the probability distribution over genera. This procedure can then be repeated to derive the distributions at all higher levels. The marginalisation loss is simply the sum of the intermediate losses computed at each level.}
\label{fig:marginalisation}
\end{figure}

Hierarchical labels derived from plant taxonomy are another source of non-visual a-priori information about plant species. The taxonomic hierarchy endows the output space with additional structure that may help to correctly classify plant species, especially if the training data is heavily imbalanced. 
Attempts to use the hierarchy rest on the assumption that closely related species in the tree have higher visual similarity than more distant ones.%
\footnote{In expectation, not necessarily in every instance}. %
On the one hand, the hierarchical grouping (for instance, of many rare species into a common genus) gives rare species statistical strength, as confusing them with each other becomes cheaper than confusing them with some frequently observed species from a different genus.
On the other hand, the grouping also benefits the fine-grained species classification, as it favours feature sharing between adjacent classes that, by themselves, have too few samples to learn a good representation \cite{tree_prior}. The taxonomic levels we use are, from the bottom to the top of the hierarchy: species, genus, family, order, class and phylum. 

%
To integrate hierarchical labels, we adopt the marginalisation loss proposed in \cite{cloth_hier}. As shown in Fig.~\ref{fig:marginalisation}, the output of the classifier is the probability distribution over all species. Marginalising over all species within each genus thus yields the probability distribution over genera. This procedure can then be repeated to derive the distribution over families, etc.:
\begin{equation}
    p(y_i^l)=\sum_{j\in K_i}p(y^{l+1}_j)
\end{equation}
where $p(y_i^l)$ is the predicted probability for the $i$-th label at hierarchy level $l$, and $p(y_j^{l+1})$ is the probability of class $j$ at the next-coarser hierarchy level $l+1$. With $K_i$ we denote the set of child classes of parent class $i$.
Based on the distribution $p(y^l)$ derived at level $l$, we can compute a cross-entropy loss $\mathcal{L}^l$ for each individual level. The marginalisation loss is then simply the sum of all these intermediate losses:

\begin{equation}
    \mathcal{L}_{\text{mar}} = \sum_l \mathcal{L}^l\;.
    \label{eq:marg}
\end{equation}

\subsection{Data Preprocessing}

All community science images were resized to the size of 256$\times$256 and then centre-cropped to 224$\times$224. The images used for training were additionally augmented by random rotations, random horizontal flips and color-jitter, which are all standard methods to help mitigate the risk of over-fitting. Furthermore, all images were normalized according to the mean and standard deviation of the training set. 

We encode the observation time, measured as day of year $t$, into $(t_1,t_2)$ using the sine-cosine mapping~\cite{geo_prior}, Eq.~\ref{eq:time}. In this way December $31^\text{st}$ and January $1^\text{st}$ are mapped close to each other, correctly accounting for the cyclic nature of the variable.

\begin{equation}
\begin{cases}
t_1 = \sin(\frac{2\pi t}{365}) \\
t_2 = \cos(\frac{2\pi t}{365}) 
\end{cases}
\label{eq:time}
\end{equation}

Regarding the location coordinates, we rescale longitude, latitude and altitude separately to fit into the interval $[-1,1]$ and denote the triple of normalised coordinates as our geo-location $(x,y,z)$.

Finally, the Sentinel-2 images are extracted from a cloud-free mosaic of images taken in 2020. As previously indicated, we only use the four spectral bands with a 10 m spatial resolution (R, G, B and N-IR), since they are often sufficient to derive vegetation parameters \cite{nico_tree}. From this mosaic, we extract patches of $256 \times 256$ pixels to ensure enough context (ca.~1.3 km around the sample location, see Tab.~\ref{tab:ablationsent2} for a comparison of the performance with different sized patches).

\subsection{Balanced Sampling}

We used a balanced sampling strategy, where the sampling weight of each image $W_i$ is inversely proportional to the number of images $N_{y_i}$ of the corresponding class $y_i$: 

\begin{equation}
    W_i = \frac{1}{N_{y_i}}~.
    \label{eq:weight}
\end{equation}

This strategy will oversample the rare species from the tail of the distribution and undersample the frequent species from the head of the distribution, so as to mitigate the impact of the imbalance on the classifier. It should be noted that the effect cannot be completely removed: even when sampled with higher frequency, the few images of a rarely observed species will inevitably carry less information than the many example images of an abundant species. 
As a result there is no clear advantage in neither of the two approaches, making this a mere design choice. In fact, using the balanced sampling strategy, compared to the conventional training method, improves the per-class accuracy while decreasing the overall accuracy (see Tab.~\ref{tab:ablationbalanced}). Although the differences in performance are small, we decide to prioritise the per-class accuracy since we believe it is more important for our application and thus choose to use the balanced sampling approach.

\begin{figure}[ht]
    \centering
    \includegraphics[width=\textwidth]{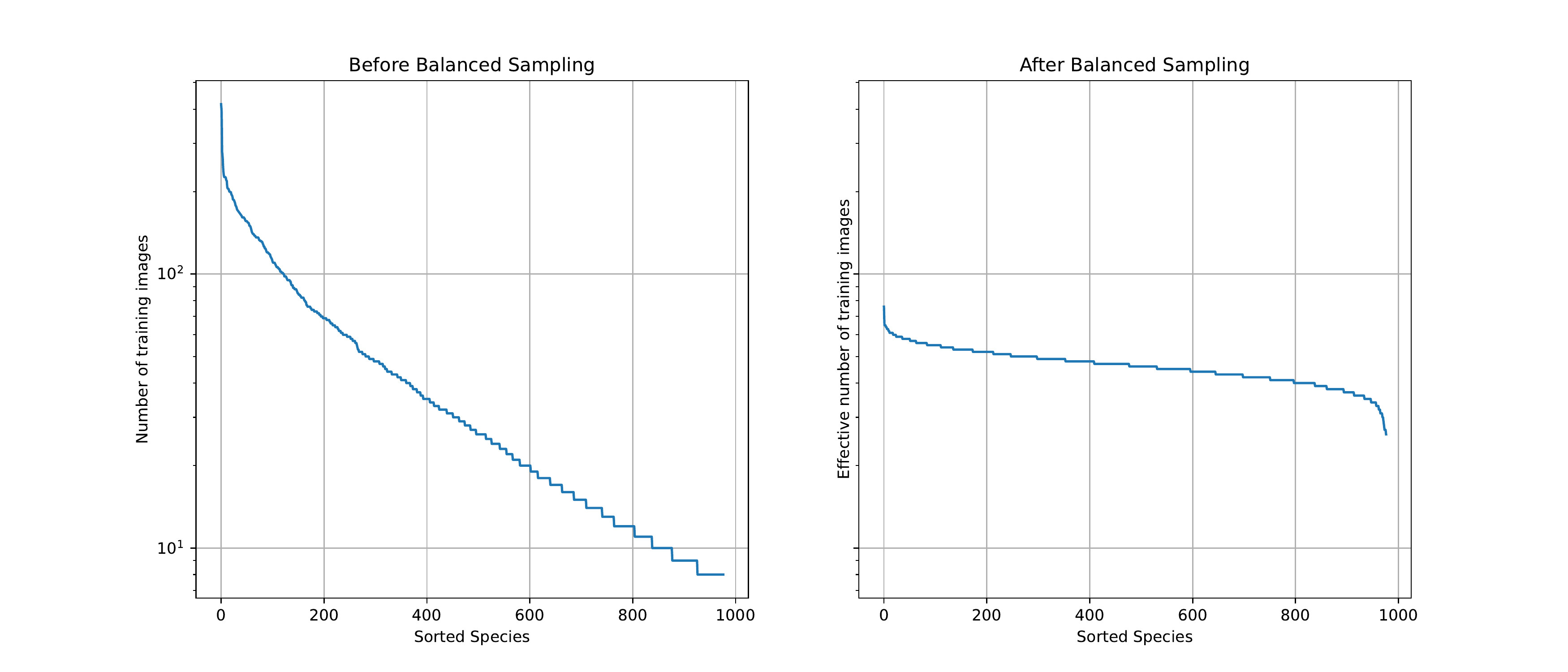}
    \caption{Sample distribution for each species in the training dataset. Note the logarithmic scale of the $y$-axis. Both diagrams share the same scale.}
    \label{fig:data_distribution}%
\end{figure}

Figure \ref{fig:data_distribution} shows the number of training images per species in the training set before and after balanced sampling.

We employ a stochastic gradient descent (SGD) \cite{sgd} to optimise the parameters of our model. We set two different learning rates, a smaller one of $5\cdot10^{-5}$ for the pre-trained convolutional layers of the CNN, and a larger one of $2\cdot10^{-3}$ for the fully connected layers. These learning rates are further reduced when a plateau is reached. The batch size is fixed to 32, and all models were trained for 100 epochs. We use a cross-entropy loss for baselines that ignore the label hierarchy, and the marginalisation loss (Eq. \ref{eq:marg}) for hierarchically structured labels. 

All results (unless stated otherwise) are computed with 5-fold cross-validation, stratified to ensure uniform class distribution across all folds.

%% file: experiments.tex
\section{Dataset}

From the iNaturalist database \cite{iNat_web}, we have downloaded all images\footnote{As of November 5th, 2020.} of plants that are located in Switzerland and labeled as ``Research Grade". The latter constitutes the highest level of data quality, where observations meet five criteria: (1) they must include a date, (2) a spatial geo-reference,  (3) a picture (or sound, but we only focus on images in this work), (4) the subject must be a naturally living organism (not captive or cultivated), and (5) at least 2 identifiers should agree on a taxon, out of a minimum of 3 identifiers.

\begin{figure}[ht]
\begin{center}
\includegraphics[width=0.65\linewidth]{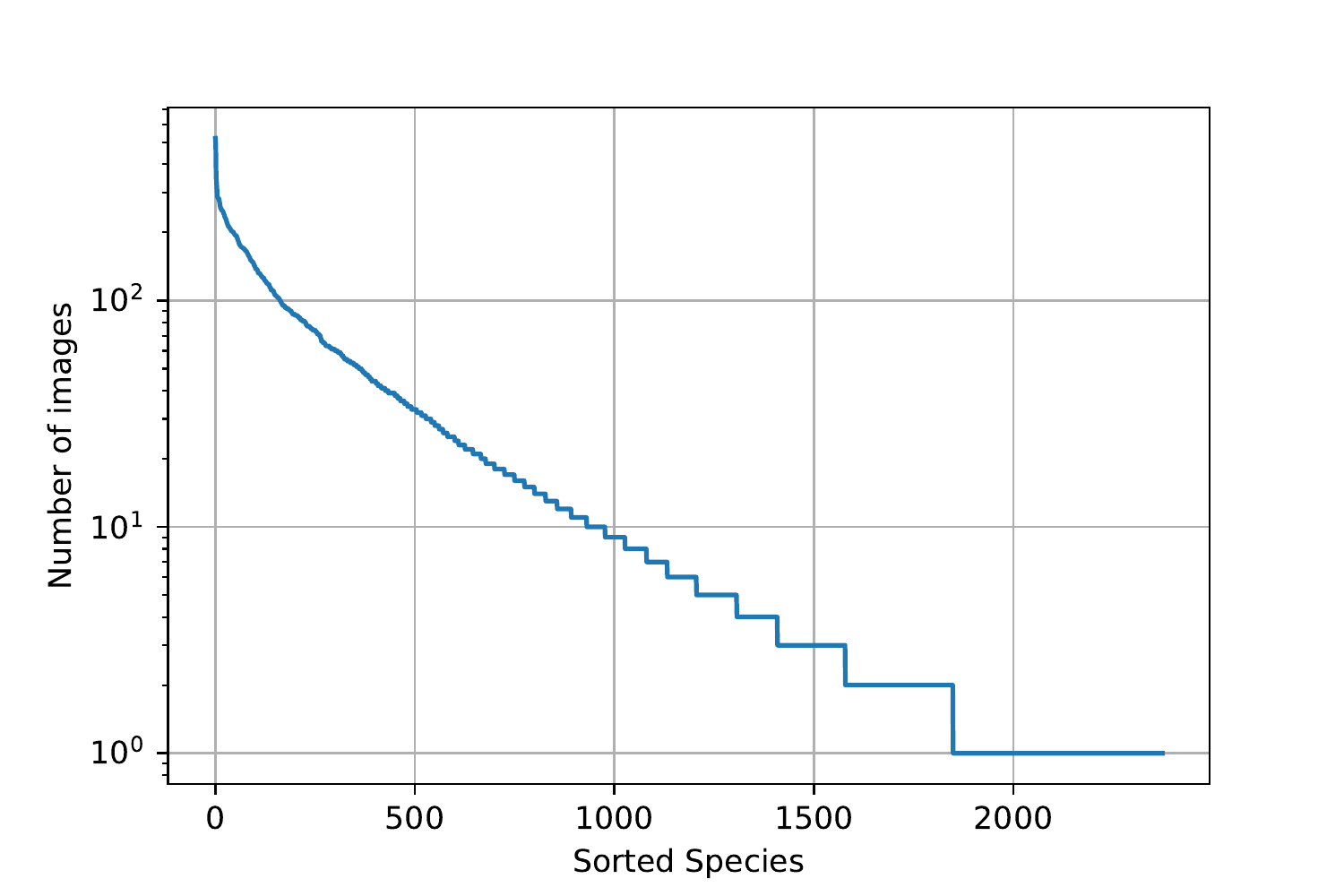} 
\end{center}
   \caption{Sample distribution for all ``Research Grade" iNaturalist observations in Switzerland. Note the logarithmic scale of the $y$-axis.}
\label{fig:overall_dist}
\end{figure}

\begin{figure}[ht]
\begin{center}
\includegraphics[width=0.85\linewidth]{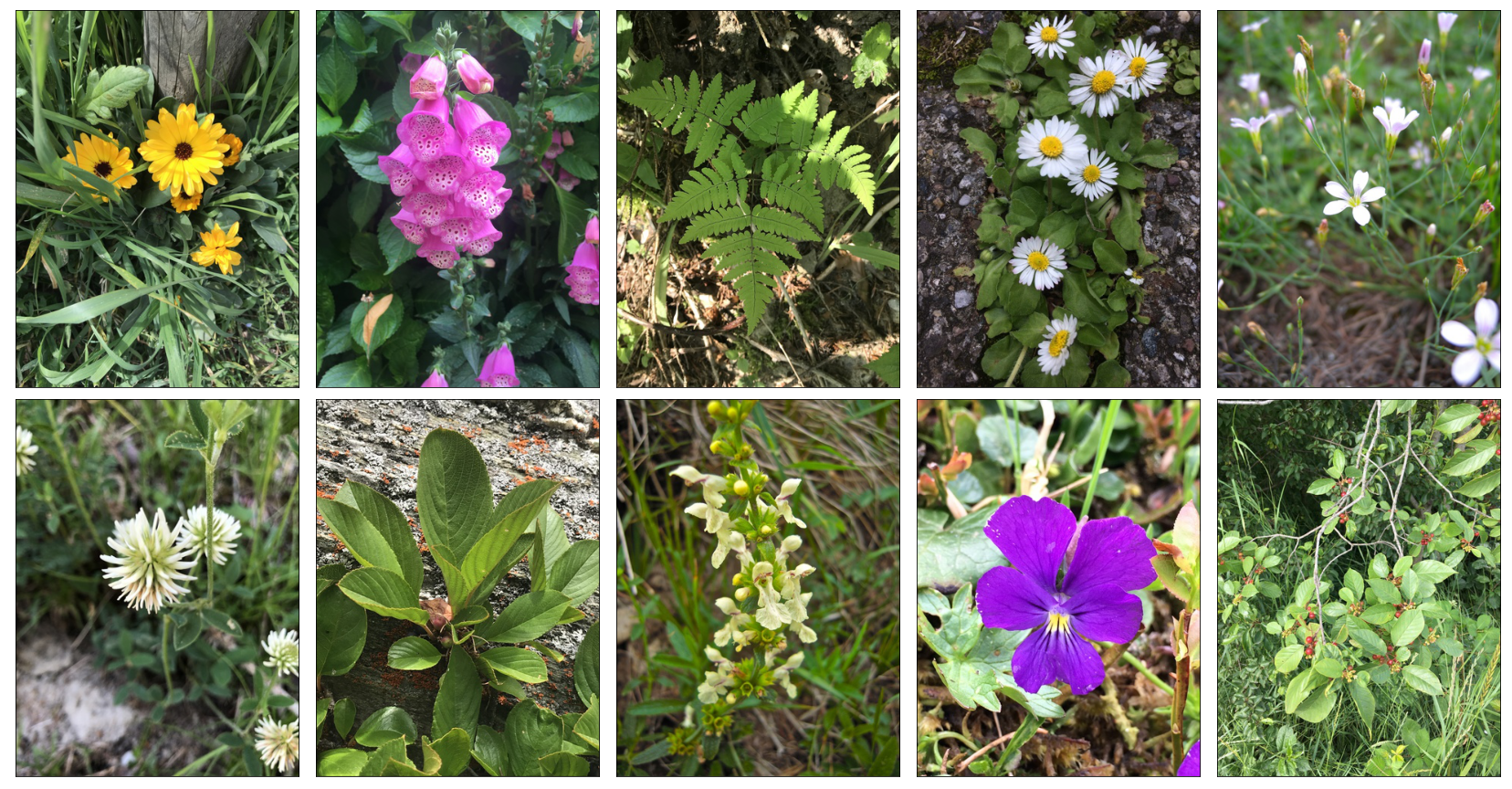} 
\end{center}
   \caption{Example images from our dataset.}
\label{fig:examples}
\end{figure}

As shown in Tab.\ref{tab:dataset}, a total of 60,781 images were downloaded (see Fig.~\ref{fig:examples}), which represented 2,374 species. However, as seen in Fig.~\ref{fig:overall_dist}, the dataset is highly imbalanced and follows a long-tail distribution. We discard all species with \textless10 images in order to ensure reliability and statistical significance of the experimental results. After this filtering we are left with 56,608 images representing 977 species. 
We also generated a dataset of unseen species for further experiments (see Sec.~\ref{sec:res_unseen}). These are observations of species that have fewer than 10 but more than 5 images. For each of those species, we select 5 images at random. 

\begin{table}[ht]
\begin{center}
\begin{tabular}{ccc}
\hline
Description & Images & Species \\
\hline
\hline
Overall & 60,781 & 2,374 \\
\hline
Selected & 56,608 & 977\\
\hline
Unseen & 1,650 & 330 \\
\hline
\end{tabular}
\end{center}
\caption{Overview of our dataset.}
\label{tab:dataset}
\end{table}

Besides the images, the dataset also contains non-visual information, including the additional data that we use in our model, i.e., longitude, latitude, day of the year and hierarchical labels. To obtain altitude we extract the height value corresponding to the given geo-location from the swissALTI3D DEM of the Swiss national mapping agency \cite{swisstopo}. 

\section{Experimental Results}

\subsection{Model Performance}

We have conducted experiments with the following models to empirically determine their performance gain: (1) \textbf{Baseline}, which corresponds to a standard ResNet50; (2) \textbf{Baseline + Location Context}, where we add the location encoder to the baseline in a late fusion setup; (3) \textbf{Baseline + Hierarchical Labels}, where we add the marginalisation loss to the baseline; and (4) \textbf{Proposed Model}, which leverages both the location context \emph{and} the hierarchical labels. Here, we again use the late fusion strategy, which empirically achieved the best performance (see Tab.~\ref{tab:training}). 

\begin{table}[ht]
\begin{center}
    \addtolength{\leftskip} {-2cm}
    \addtolength{\rightskip}{-2cm}
\begin{tabular}{|l|c|c|c|c|}
\hline
Model & Accuracy (\%) & Top-1 (\%) & Top-3 (\%) & Top-5 (\%) \\
\hline\hline
Baseline &  73.48 & 62.48 & 79.04 & 83.97 \\
\hline
Baseline + Location Context & 76.99 & 67.47 & 82.50 & 87.01 \\
\hline
Baseline + Hierarchical Labels & 76.30 & 65.49 & 82.20 & 86.81 \\
\hline
\textbf{Proposed Model} & \textbf{79.12} & \textbf{69.76} & \textbf{84.86} & \textbf{88.95} \\
\hline
\end{tabular}
\end{center}
\caption{Ablation study of our proposed model. Note that the top-k accuracies denote the average species-specific metrics in order to give the same weight to all the species in the evaluation.}
\label{tab:ablation}
\end{table}

\begin{figure}[ht]
\begin{center}
\includegraphics[width=.85\linewidth]{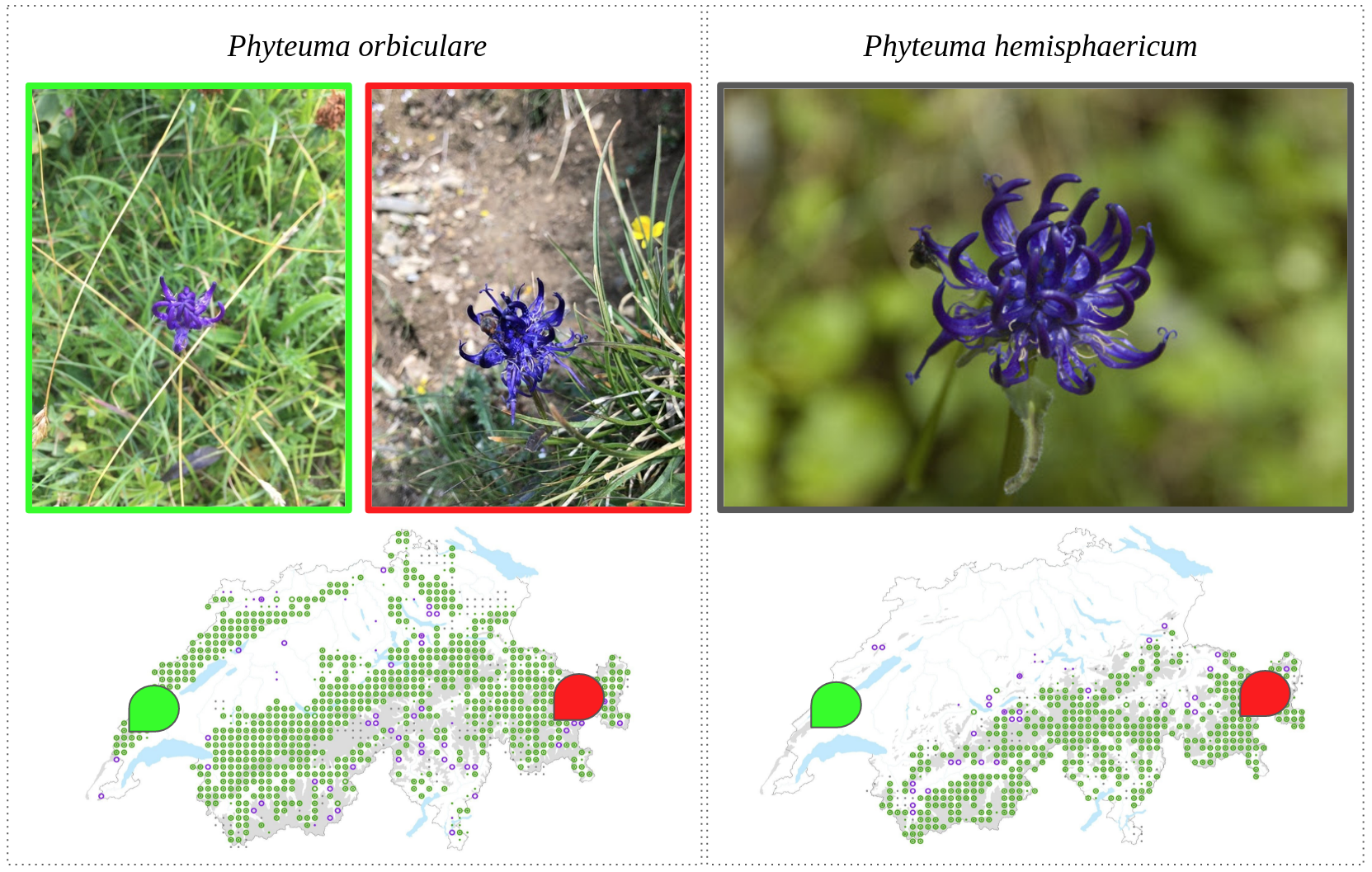} 
\end{center}
   \caption{Misclassification example: both images of \textit{Phyteuma orbiculare} (Left) are misclassified as \textit{Phyteuma hemisphaericum} (Right) by our baseline model. When including the location context, our proposed model correctly classifies the image with the green frame, whereas the image with the red frame is still misclassified. The green and red arrows indicate the locations of the respective left two images. The underlying maps are the species distribution maps downloaded from InfoFlora \cite{infoflora_web}. This highlights the importance of including the location information to distinguish visually similar species that have different geographical ranges.}
\label{fig:misclassification}
\end{figure}

As seen in Tab.~\ref{tab:ablation}, adding either the location context or the hierarchical labels to the baseline model significantly improves the results, for all metrics. Note that we compute the top-k accuracies as the average species-specific metrics in order to give the same weight to all the species in the evaluation. Thus the overall accuracy is higher than the top-1 hit-rate due to the imbalanced nature of our dataset, which is preserved in our stratified cross-validation. Furthermore, improvements from location context and hierarchical labels are largely orthogonal, as expected, since they leverage different types of information. These results indicate a clear benefit of complementing visual cues from community science images with additional sources of information. For a more detailed ablation study of the exact contributions of every component in our location context see Sec.~\ref{sec:context}. Visual inspection of misclassified images confirms that location context helps in the case of visually similar species that occur in different geographical regions (see Fig.~\ref{fig:misclassification}), whereas hierarchical labels help to classify species with few images. 

\begin{figure}[ht]
\begin{center}
\includegraphics[width=.85\linewidth]{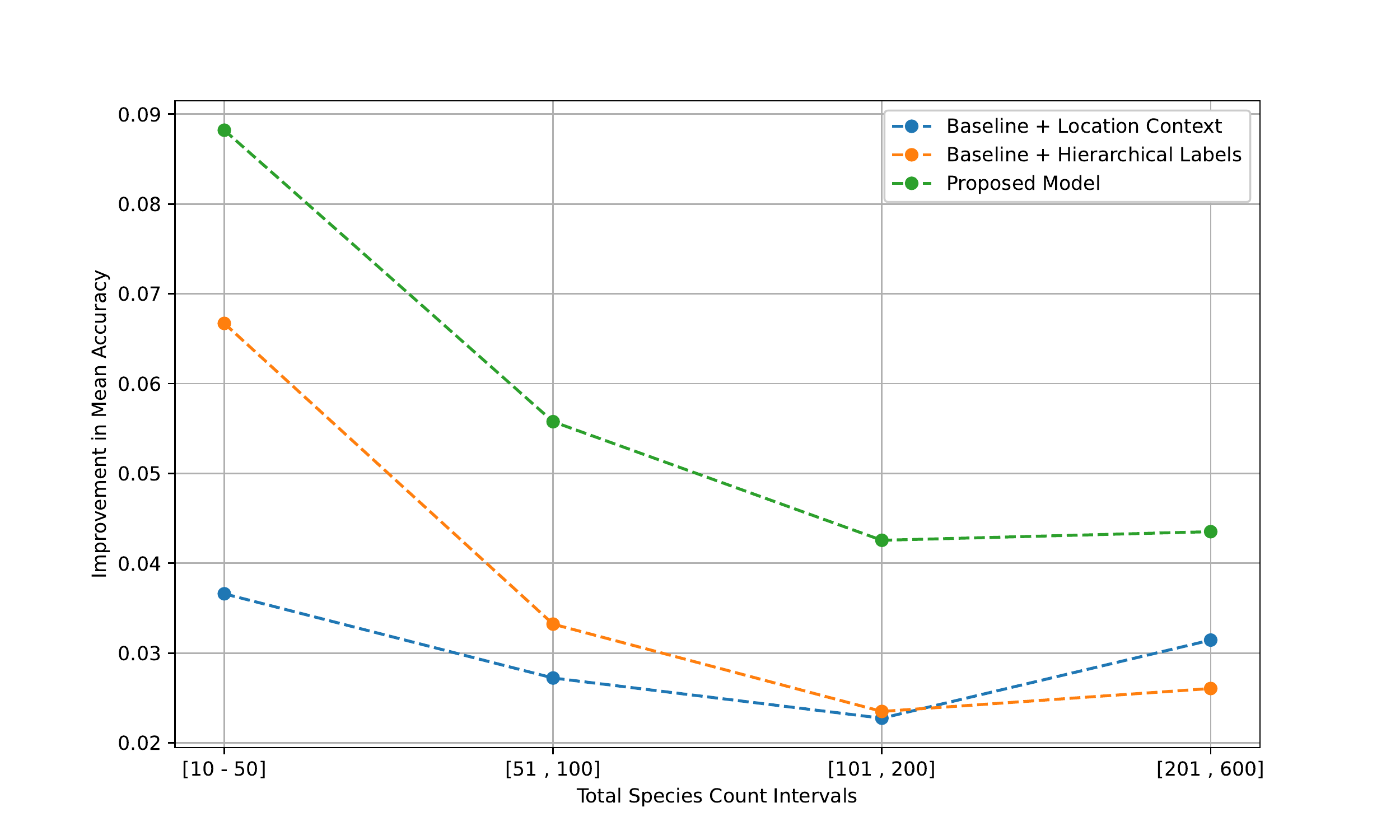} 
\end{center}
   \caption{Improvement in Mean Accuracy over the baseline for species with different numbers of images in the dataset.}
\label{fig:improvement}
\end{figure}

Finally, as seen in Fig.~\ref{fig:improvement} our proposed model improves over the baseline for all four ranges of species counts and the margin of improvement is largest for the tail species of the dataset with a number of images between 10 and 50. This is very useful since rare species are more commonly misidentified by community science and are particularly important for conservation purposes. 

\newpage
\subsection{Training Strategies}

\begin{table}[ht]
\begin{center}
    \addtolength{\leftskip} {-2cm}
    \addtolength{\rightskip}{-2cm}
\begin{tabular}{|l|c|c|c|c|}
\hline
Model & Accuracy (\%) & Top-1 (\%) & Top-3 (\%) & Top-5 (\%) \\
\hline\hline
Separate Training & 76.41 & 65.29 & 82.09 & 86.58 \\
\hline
Joint Training: Early Fusion & 73.65 & 64.84 & 79.60 & 83.87 \\
\hline
\textbf{Joint Training: Late Fusion} & \textbf{79.12} & \textbf{69.76} & \textbf{84.86} & \textbf{88.95} \\
\hline
\end{tabular}
\end{center}
\caption{Comparison of different training strategies. Note that the top-k accuracies indicate the average species-specific metrics.}
\label{tab:training}
\end{table}

Table~\ref{tab:training} compares the three different training strategies described in Sec.~\ref{sec:train_strat}. The separate training strategy has the advantage that one can use the image classifier and get reasonable predictions even when metadata is missing. Whereas the joint training strategies should always perform better, at the cost of being less flexible, as metadata is mandatory. Under ideal circumstances, one would also expect the early fusion strategy to perform best, as it is not subject to any factorisation constraints on $p(y|I,\mathbf{x})$ and can leverage the complete correlation structure. In practice, we however observe the worst performance, see Tab.~\ref{tab:training}. It appears that the increased model capacity leads to over-fitting. The late fusion training strategy, with its restricted interaction between image and context cues, emerges as the best compromise with clearly superior performance. Separate training does bring a noticeable improvement over the baseline but does not reach the late fusion approach. Likely this is, at least in part, due to the presence-only labels hampering the learning of the prior $p(y|\mathbf{x})$.

\subsection{Evaluation at Different Hierarchy Levels}

When using the taxonomic hierarchy during training in conjunction with the marginalisation loss, we can predict at inference time labels at different hierarchy levels. If taxonomic distance indeed correlates with similar visual features and ecological requirements (see Fig.~\ref{fig:confusion}), then the predictions at higher levels should be increasingly more correct. I.e., even if a specimen is assigned the wrong species label it might be assigned the correct genus label, as it is more likely to be confused with a similar species from the same genus.%
\footnote{Note that also the chance level increases, as there are fewer possible labels.}

We have evaluated our model at all taxonomic levels that we use, see Tab.~\ref{tab:hier_res}. Indeed, the performance is better for the higher levels (c.f.~Tab.~\ref{tab:hier_dataset}).
Furthermore, higher up in the hierarchy, fewer classes are poorly represented; the long-tail distribution is less extreme.

\begin{table}[ht]
\begin{center}
\begin{tabular}{ccccccc}
\hline
Level & Species & Genus & Family & Order & Class & Phylum \\
\hline
\hline
Number & 977 & 489 & 121 & 50 & 8 & 3 \\
\hline
\end{tabular}
\end{center}
\caption{Number of classes at each hierarchical level.}
\label{tab:hier_dataset}
\end{table}

\begin{table}[h]
\begin{center}
\begin{tabular}{|l|c|c|c|c|c|c|}
\hline
Metric (\%) & Species & Genus & Family & Order & Class & Phylum \\
\hline\hline
Accuracy & 79.02 & 83.39 & 87.26 & 88.54 & 97.24 & 99.89 \\
\hline
Top-1 & 69.50 & 73.23 & 75.84 & 78.53 & 88.52 & 86.63 \\
\hline
Top-3 & 84.49 & 85.76 & 87.81 & 89.34 & 98.01 & 100.0 \\
\hline
Top-5 & 88.57 & 89.37 & 91.45 & 92.95 & 99.51 & 100.0 \\
\hline
\end{tabular}
\end{center}
\caption{Results at different hierarchical levels. Note that top-3 and top-5 accuracy at phylum level are meaningless, since there are only 3 possible phyla. Note that the top-k accuracies indicate the average species-specific metrics.}
\label{tab:hier_res}
\end{table}

\newpage
\subsection{Experiments with Unseen Species}
\label{sec:res_unseen}

Given the hierarchical labels, it is also possible to classify new species which the classifier has not seen at all during training. While the assigned species label will necessarily always be wrong, one would hope that the predictions at coarser taxonomy levels are often sensible.
For this experiment, we picked 330 species that were initially discarded from our dataset for having \textless10 images, but for which at least 5 images are available, c.f.\ the "Unseen" row in Tab.~\ref{tab:dataset}. The corresponding results in Tab.~\ref{tab:unseen} confirm our intuition: while there is of course a significant performance drop compared to the trained species, it is still possible to classify unseen species into the right Genus, Family or Order with reasonable performance, well above chance level (the probability of success of a classifier that always predicts the most common class). 
This capability can be extremely useful in the context of community science, where the coarser labels can be used to refer examples to the right expert for classification or to detect gaps in the taxonomy lists offered to users.

\begin{table}[ht]
\begin{center}
\begin{tabular}{|l|c|c|c|c|c|c|}
\hline
Evaluation Set & Species & Genus & Family & Order & Class & Phylum \\
\hline\hline
5-fold Cross-Val &  79.02 & 83.39 & 87.26 & 88.54 & 97.24 & 99.89 \\
\hline
Unseen Species & - & 24.27 & 41.86 & 50.23 & 85.60 & 96.00 \\
\hline
\end{tabular}
\end{center}
\caption{Accuracy (\%) on unseen species.}
\label{tab:unseen}
\end{table}

\subsection{Contextual Information and Sentinel-2 Images Ablation Study}

\label{sec:context}

To investigate the contributions of different types of contextual information, and the potential benefit of adding satellite imagery, we perform extensive ablation studies.

In Tab.~\ref{tab:ablationspatiotemporal} we show the impact of the different contextual information (altitude, geo-coordinates, day of the year) on the evaluated metrics. As it can be seen they all contribute to some extent, with the altitude being one of the most important. Considering the high altitude variability of the Swiss landscape it was rather expected that the altitude could carry the most valuable information. When the full context is combined the performance metrics show a further increase meaning that the additional data carry orthogonal information.

Finally, Tab.~\ref{tab:sent2} displays the performance achieved with the integration of Sentinel-2 imagery. Overall, their impact turns out to be small. When naively adding the Sentinel-2 branch, performance even drops slightly, apparently due to over-fitting.
By adding standard drop-out regularisation~\cite{dropout} on the last fully-connected layer, we were able to remedy this behaviour and achieve a mild (but still statistically significant) performance gain.
To ensure that the difference is actually caused by the satellite imagery and not the drop-out, we add an additional baseline where the model without the Sentinel-2 branch is trained with drop-out. Interestingly, this even degraded the performance.

While it is promising that the much-enriched context information from the satellite image brings an improvement over the simple geo-location, that gain is relatively modest, at least with our implementation. Further research, beyond the scope of the present paper, will be needed to clarify the potential of satellite (or airborne) data as auxiliary information. 

\begin{table}[h]
\begin{center}
    \addtolength{\leftskip} {-2cm}
    \addtolength{\rightskip}{-2cm}
\begin{tabular}{|l|c|c|c|c|}
\hline
Model & Accuracy (\%) & Top-1 (\%) & Top-3 (\%) & Top-5 (\%) \\
\hline\hline
Baseline &  73.48 & 62.48 & 79.04 & 83.97 \\
\hline
Baseline + Altitude & 75.40 & 65.11 & 81.00 & 85.80 \\
\hline
Baseline + Geo-coordinates & 75.07 & 64.86 & 80.63 & 85.45 \\
\hline
Baseline + Day of the year & 75.51 & 65.00 & 80.91 & 85.51 \\
\hline
\textbf{Baseline + Full Location Context} & \textbf{76.99} & \textbf{67.47} & \textbf{82.50} & \textbf{87.01} \\
\hline
\end{tabular}
\end{center}
\caption{Ablation study of spatio-temporal context. Note that the top-k accuracies indicate the average species-specific metrics.}
\label{tab:ablationspatiotemporal}
\end{table}

\begin{table}[h]
\begin{center}
    \addtolength{\leftskip} {-2cm}
    \addtolength{\rightskip}{-2cm}
\begin{tabular}{|l|c|c|c|c|}
\hline
Model & Accuracy (\%) & Top-1 (\%) & Top-3 (\%) & Top-5 (\%) \\
\hline\hline
Proposed model & 79.12 & 69.76 & 84.86 & 88.95 \\
\hline
Proposed model with Dropout & 78.02 & 67.77 & 83.39 & 87.52 \\
\hline
Proposed model + Sen-2 & 78.59 & 68.29 & 84.43 & 88.60 \\
\hline
\textbf{Proposed model + Sen-2 with Dropout} & \textbf{79.73} & \textbf{70.32} & \textbf{85.52} & \textbf{89.33} \\
\hline
\end{tabular}
\end{center}
\caption{Results adding Sentinel-2 mosaic. Note that the top-k accuracies indicate the average species-specific metrics.}
\label{tab:sent2}
\end{table}

%% file: conclusion.tex
\section{Conclusion}

In this work, we have demonstrated that easily accessible side information can bring rather large performance gains when classifying community science photographs. 
We have focused on the spatio-temporal context of the observations,
and have shown how it can refine the classification model by providing relevant prior knowledge regarding the distribution and occurrence of species observations.
We have also briefly touched on extended radiometric context from optical satellite imagery, a direction where we see quite some potential for further research.
Moreover, we have verified that exploiting the hierarchical structure of biological taxonomy not only improves the species recognition performance, but also enables more reliable predictions at coarser taxonomy levels, and even coarse classification of species not seen at all during the classifier training.
  
In terms of practical community science applications, our model is also a step towards a viable scheme for verifying user-supplied labels. For instance, the proposed method could provide hints to the community scientist when labelling the species, or it could facilitate the reviewing validation by experts, marking specific observations where the model disagrees with the label provided by the community scientist. Of course, these suggestions would need to be followed with care in practice to avoid creating a confirmation bias of the model.
We hope that, ultimately, a larger number of correct species observations will contribute to better species distribution models, to inform biodiversity research and conservation initiatives, particularly for rare species.

%% file: appendix.tex
\newpage
\appendix

\renewcommand{\thefigure}{A\arabic{figure}}
\setcounter{figure}{0}
\setcounter{table}{0}

\section{Ablation Studies}

\begin{table}[ht]
\begin{center}
    \addtolength{\leftskip} {-2cm}
    \addtolength{\rightskip}{-2cm}
\begin{tabular}{|l|c|c|c|c|}
\hline
Model & Accuracy (\%) & Top-1 (\%) & Top-3 (\%) & Top-5 (\%) \\
\hline\hline
Baseline + No Balanced Sampling & 75.57 & 62.08 & 80.9 & 86.18
\\
\hline
Baseline + Balanced Sampling & 73.48 & 62.48 & 79.04 & 83.97 \\
\hline
Proposed Model + No Balanced Sampling & \textbf{80.05} & 69.23 & \textbf{86.15} & \textbf{90.29} \\
\hline
Proposed Model + Balanced Sampling & 79.12 & \textbf{69.76} & 84.86 & 88.95 \\
\hline
\end{tabular}
\end{center}
\caption{Ablation study of balanced sampling. Note that the top-k accuracies denote the average species-specific metrics.}
\label{tab:ablationbalanced}
\end{table}

\begin{table}[ht]
\begin{center}
    \addtolength{\leftskip} {-3cm}
    \addtolength{\rightskip}{-3cm}
\begin{tabular}{|l|c|c|c|c|}
\hline
Model & Accuracy (\%) & Top-1 (\%) & Top-3 (\%) & Top-5 (\%) \\
\hline\hline
No Sent-2 & 79.12 & 69.76 & 84.86 & 88.95 \\
\hline
Small Sent-2 ($128\times128$)& 79.5 & 69.56 & 84.76 & 88.92 \\
\hline
\textbf{Normal Sent-2 ($256\times256$)} & \textbf{79.73} & \textbf{70.32} & \textbf{85.52} & \textbf{89.33} \\
\hline
Large Sent-2 ($512\times512$) & 79.16 & 69.9 & 84.94 & 88.91 \\
\hline
\end{tabular}
\end{center}
\caption{Comparison of different extents for Sentinel-2 images. Note that the top-k accuracies indicate the average species-specific metrics.}
\label{tab:ablationsent2}
\end{table}

\newpage
\section{Confusion Matrix}

\begin{figure}[ht]
\begin{center}
\includegraphics[width=.95\linewidth]{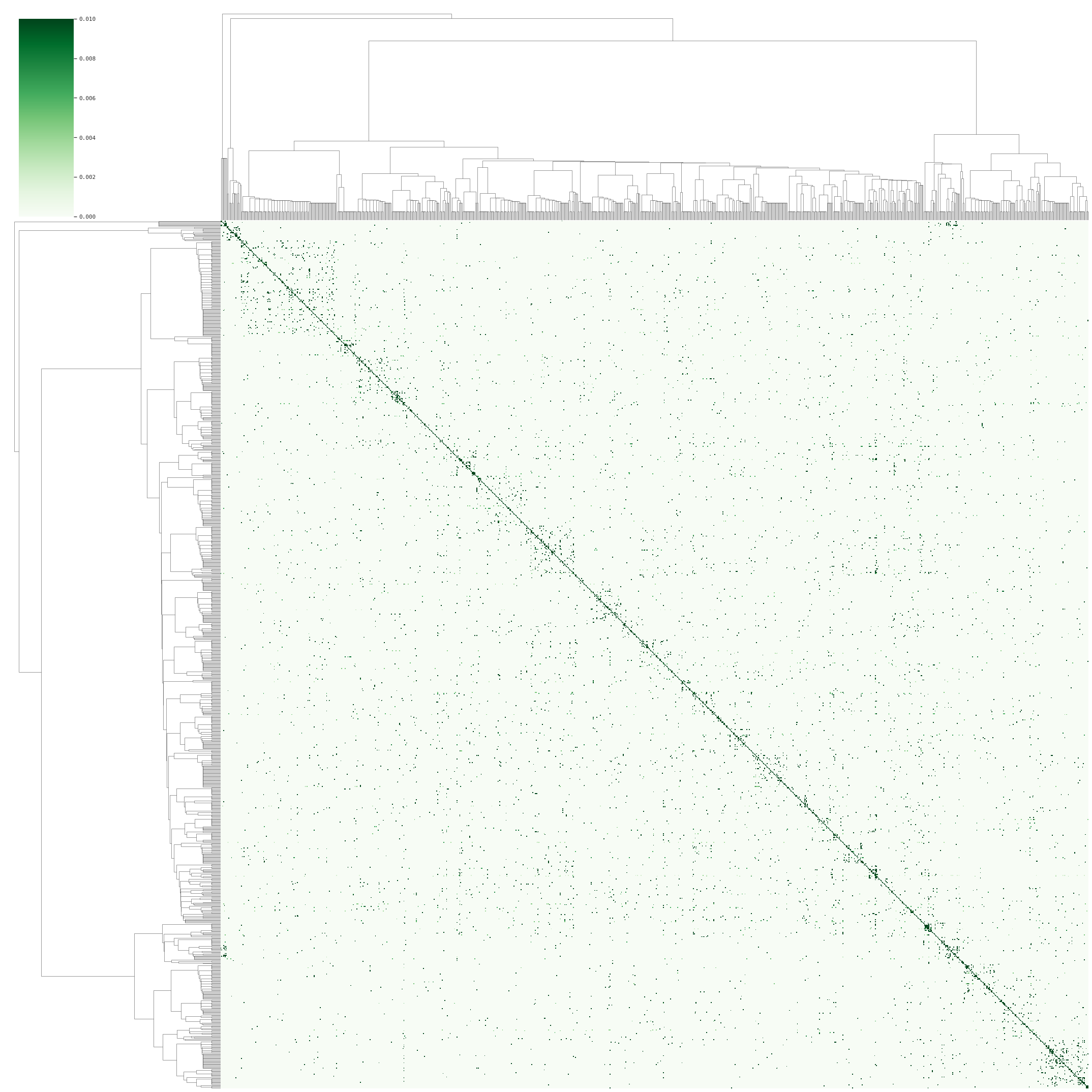} 
\end{center}
   \caption{Confusion matrix for our proposed model. The species in the rows and columns have been ordered based on their taxonomy. The hierarchy between the species is made apparent by the dendrograms. The block-like structure along the diagonal indicates that species that are close in terms of their taxonomy are misclassified for each other more often than unrelated species.}
\label{fig:confusion}
\end{figure}